\documentclass[12pt]{article}
 \usepackage{amsmath}
\usepackage{fancyhdr}
\usepackage{graphicx}
\usepackage{color}
\usepackage{dsfont}
\usepackage{stmaryrd} 
\usepackage{booktabs}
\usepackage{wrapfig}
\usepackage[left=2.5cm,right=2.5cm,top=2.5cm,bottom=2.5cm]{geometry}
\begin{document}
\renewcommand{\headrulewidth}{0pt}
\fancyfoot[L]{ JdS2025}
\fancyhead[R]{ }


\begin{center}
{\Large
	{\sc Une approche policy-gradient pour des actions ordinales
}
}
\bigskip

\underline{Simon Weinberger} $^{1, 2}$ \& Jairo Cugliari $^{2}$
\bigskip

{\it
$^{1}$ Essilor International, affiliate of EssilorLuxottica\\
$^{2}$ Laboratoire ERIC, Université Lumière Lyon 2, ERIC, \texttt{jairo.cugliari@univ-lyon2.fr}
}
\end{center}
\bigskip


{\bf R\'esum\'e.} En apprentissage par renforcement, la paramétrisation softmax est l’approche standard pour les politiques sur des espaces d’actions discrètes. Cependant, elle ne prend pas en compte la relation d’ordre entre les actions. Motivés par un problème industriel réel, nous proposons une nouvelle paramétrisation de politiques basée sur les modèles de régression ordinale, adaptés au cadre de l’apprentissage par renforcement. Notre approche répond aux défis pratiques, et des expériences numériques démontrent son efficacité dans des applications réelles et dans des tâches à actions continues, où la discrétisation de l’espace d’actions combinée à la politique ordinale offre des performances compétitives.

{\bf Mots-cl\'es.} Apprentissage par renforcement, Policy-gradient, Régression ordinale. 

\medskip

{\bf Abstract.} In reinforcement learning, the softmax parametrization is the standard approach for policies over discrete action spaces. However, it fails to capture the order relationship between actions. Motivated by a real-world industrial problem, we propose a novel policy parametrization based on ordinal regression models adapted to the reinforcement learning setting. Our approach addresses practical challenges, and numerical experiments demonstrate its effectiveness in real applications and in continuous action tasks, where discretizing the action space and applying the ordinal policy yields competitive performance.

{\bf Keywords.} Reinforcement Learning, Policy-gradient, Ordinal Regresion.

\bigskip\bigskip


\section{Introduction}

L'apprentissage par renforcement (\textsf{RL}) permet de résoudre des problèmes nécessitant des décisions séquentielles afin de maximiser une récompense à long terme, suivant une politique à déterminer. Des algorithmes tels que \textit{Trust Region Policy Optimization} (\textsf{TRPO}, Schulman \textit{et al.}, 2015), \textit{Proximal Policy Optimization} (\textsf{PPO}, Schulman \textit{et al.}, 2017) ou \textit{Natural Policy Gradient} (\textsf{NPG}, Kakade, 2001) estiment des politiques paramétrées qui induisent une distribution sur l'espace d'actions. Toutefois, la majorité des applications reposent sur une distribution logistique (\textit{softmax}) pour les actions catégoriques (Agarwal \textit{et al.}, 2020) ou une distribution normale pour les actions continues (Sutton et Barto, 2018), négligeant la relation d'ordre entre les actions ordinales.

Cette limitation affecte des problèmes réels où les actions suivent un ordre naturel. Par exemple, des verres électrochromiques d’EssilorLuxottica, qui proposent quatre niveaux de teinte ordonnés ajustés en fonction de la lumière ambiante via un capteur (\textit{Ambient Light Sensor}, \textsf{ALS}) et des seuils prédéfinis. 
Nous proposons un algorithme de contrôle de teinte pour des lunettes électrochromiques, ajustant la teinte parmi des classes ordonnées allant d'une classe claire (i.e. verres transparents) à la clase plus obscure (des lunettes de soleil foncées). 

Dans un contexte de \textsf{RL}, le tâche à résoudre implique des capteurs, qui mesurent l’état, une proposition d'une teinte (action), et l'acceptation de la teinte par l’utilisateur ou un choix manuel alternatif (récompense). L’objectif est d’optimiser la politique pour maximiser les récompenses actualisées, tout en tenant compte des éventuels délais de réaction. Ce processus est itéré pendant un épisode (par exemple, une journée), permettant des mises à jour dynamiques du contrôle de la teinte.

Nous proposons une politique ordinale inspirée des modèles de régression ordinale linéaire (Agresti, 2010) et des réseaux de neurones CORAL (Cao \textit{et al.}, 2020). Cette approche permet l’utilisation de méthodes d’optimisation par gradient de politique telles que \textsf{REINFORCE}, \textsf{NPG}, \textsf{TRPO} et \textsf{PPO}. Nos résultats expérimentaux montrent que, dans des contextes où les actions sont ordonnées, les politiques ordinales convergent plus rapidement et vers de meilleures solutions que les politiques multinomiales classiques (i.e.~\textit{softmax}). De plus, elles offrent des performances comparables à celles des politiques continues lorsque l’espace d’actions continues est discrétisé de manière appropriée.


\section{Cadre théorique}

Nous considérons un processus de décision markovien où 
$\mathcal{S}$, $\mathcal{A}$ et $\mathcal{R}=\mathds{R}$ représentent les espaces d’états, d’actions et de récompenses, respectivement. L'évolution du système est régie par une densité (inconnue) de transition d’état :
\[
p(s', r | s, a) := \mathds{P}(S_{t+1}=s', R_{t+1}=r | S_t = s, A_t = a), \qquad s, s' \in\mathcal{S}, a \in\mathcal{A}, r \in\mathcal{R}.
\]
À chaque moment (ou étape) $t$, les actions sont prises selon une politique, $\pi(.|s_t)$, que nous considérons étant une distribution de probabilité paramétrique, c'est-à-dire nous posons $\pi = \pi_{\theta}$ où le vecteur de paramètres $\theta \in \Theta \subset \mathds{R}^d$.
Pour un facteur d’actualisation donné $\gamma \in [0,1[$, une action $a \in \mathcal{A}$, un état $s \in \mathcal{S}$ et une politique $\pi$, nous définissons la fonction de valeur d’état-action $Q^{\pi}(s,a)$ et la fonction de valeur d’état $V^{\pi}(s)$ comme étant respectivement :
\begin{align*}
    Q^{\pi}(s,a) = \mathds{E}_{\pi_{\theta}}\left( \sum_{t=0}^\infty \gamma^t R_t | S_0 = s, A_0=a \right) \quad \text{et} \quad
    V^{\pi}(s) = \mathds{E}_{\pi_{\theta}}\left( \sum_{t=0}^\infty \gamma^t R_t | S_0 = s \right).
\end{align*}
L'objectif de l'optimisation de politique est de trouver $\theta \in \Theta$ qui maximise la fonction objective $J$, définie comme :
\[
J(\theta) = \mathds{E}_{\pi_{\theta}}\left( \sum_{t=0}^\infty \gamma^t R_t \right).
\]
%

\section{Politiques ordinales}


Nous considérons l'espace des actions $\mathcal{A} $ un ensemble fini d'éléments 
$ \{a_k : k = 1, \ldots, K\}$, où $K$ est un nombre naturel supérieur ou égal à 3, et il existe une relation d'ordre entre les actions : $a_1 < a_2  < \ldots < a_{K-1} < a_K$. Sans perte de généralité, nous identifions les éléments avec leurs étiquettes $\mathcal{A} =\llbracket 1, K\rrbracket$. 

\paragraph{Paramétrisation} Soit une fonction $g_{\omega}\colon \mathcal{S}\to\mathds{R}$ paramétrée par un vecteur de poids $\omega \in \Omega$, un ensemble de $K-1$ seuils ordonnés 
$\tau_1 < \tau_2 <\ldots < \tau_{K-2} < \tau_{K-1}$ et la fonction sigmoïde $\sigma(x) = (1+\exp(-x))^{-1}$. Nous vodrions une politique qui choisira l'action $A = j$ si la quantité latente $A^*= g_{\omega}(s) + e, e\sim \text{Logistique}(0,1)$, se trouve entre les seuils $ \tau_{j-1} < A^* \leq \tau_j$. Cette formulation classique en régression ordinale est équivalente à l'expression 
\begin{equation}\label{eq:ordinal_def}
    \mathds{P}(A \leq j| S=s)  = \sigma \left( \tau_j - g_{\omega}(s)\right).
\end{equation}
En utilisant la convention $\tau_0=-\infty$ et $\tau_K = \infty$, l’équation (\ref{eq:ordinal_def}) induit une distribution de probabilité sur l’espace des actions $\mathcal{A}$ et qu'on appellera \textit{politique ordinale}, définie par 
$$\pi(a|s) = \sigma(\tau_a - g_{\omega}(s)) - \sigma(\tau_{a-1} - g_{\omega}(s)), a\in\mathcal{A}.$$ 
%

Intuitivement, la fonction $g_{\omega}$ est une forme de score, plus elle est élevée plus la politique choisira actions élevées.
%
L'espace paramétrique de la politique ordinale est $\Theta = \Omega\times \Delta_{K-1}$ où
$\Delta_{K-1}$ est l'ensemble de seuils ordonnées.

\section{Expériences numériques}




\paragraph{Données simulées} La mesure \textsf{ALS} est simulée par un processus gaussien. La réponse utilisateur est modélisée par une politique ordinale inconnue $\pi_U$. À un instant donné, une classe $a_t$ est proposée pour l’état $s_t$, et un score de désagrément $Z_t$ est mis à jour selon :
\[
Z_{t+1} = (1-\pi_{U}(a_t|s_t))^{\gamma_{r}} + \gamma_{d} Z_t.
\]
L’utilisateur réagit avec une probabilité $\sigma(Z_{t+1})$ et, le cas échéant, sélectionne une classe selon $\pi_U(.|s_t)$. Le paramètre $\gamma_{r}\in\mathds{R}^+$ contrôle la fréquence des réactions : une valeur faible entraîne des ajustements fréquents, tandis qu’une valeur élevée signifie que l’utilisateur n’intervient que lorsque la classe proposée est improbable sous $\pi_U$. Le paramètre $\gamma_{d}\in[0,1]$ modélise la mémoire de l’utilisateur, une valeur nulle indiquant une réaction uniquement basée sur la classe actuelle.

\paragraph{Apprentissage} L’espace des états $\mathcal{S}$ correspond aux mesures du \textsf{ALS}, l’espace des actions $\mathcal{A}$ aux quatre classes de teinte, et l’espace des récompenses $\mathcal{R} = \{-3, -2, -1, 0\}$ à la différence absolue entre la teinte proposée et celle choisie par le porteur. Nous comparons les politiques ordinales et softmax mises à jour avec \textsf{REINFORCE}, \textsf{NPG} et \textsf{TRPO}. Nous utilisons $\gamma_{r}= 0.5$, $\gamma_{d} = 1$ et un taux d’actualisation de $0.9$. Chaque épisode comprend soixante étapes et se conclut par une mise à jour de la politique. Pour chaque combinaison de stratégie de mise à jour et de politique, nous simulons quatre cents épisodes, générant ainsi dix trajectoires d’apprentissage par configuration.


\begin{figure}[h]
    \centering
    \includegraphics[width=0.7\linewidth]{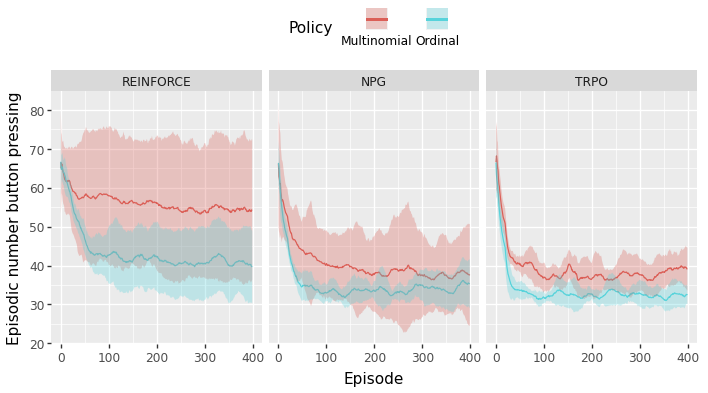}
    \caption{Récompense totale par épisode en fonction du nombre d’épisodes, pour les méthodes \textsf{REINFORCE}, \textsf{NPG} et \textsf{TRPO} et les politiques ordinale et multinomial. Les courbes moyennes sont calculées sur dix tirages aléatoires, avec un lissage par moyenne glissante sur une fenêtre de vingt épisodes. Les zones ombrées représentent la moyenne $\pm$ un écart-type.}
    \label{fig:summary_ordinal_vs_mn}
\end{figure}

\paragraph{Résultats} Les politiques ordinales convergent plus rapidement et vers de meilleures solutions que les politiques multinomiales. De plus, cette paramétrisation se révèle plus stable, comme en témoigne la diminution de l’écart-type lors de l’utilisation d’une politique ordinale (voir Figure \ref{fig:summary_ordinal_vs_mn}). Parmi les six méthodes testées, la mise à jour d’une politique ordinale avec \textsf{TRPO} produit les meilleurs résultats, offrant une amélioration plus rapide que \textsf{REINFORCE} et une stabilité légèrement supérieure à celle de \textsf{NPG}.

\paragraph{Discrétisation d’un espace d’actions continues}\label{sec:discretization}

Les actions ordinales peuvent émerger de la discrétisation d’un espace d’actions continues. Considérons un espace borné $\mathcal{A}=[m,M]$, que nous discrétisons selon : $\mathcal{A}_K = \left\{ a_k \right\}_{k=1}^K$ avec $a_k = m + k \frac{M-m}{K}.$ L’ensemble $\mathcal{A}_K$ est naturellement ordonné et possède un nombre fini d’actions. 

Dans le cas d’actions continues multivariées, la discrétisation peut être appliquée séparément à chaque dimension. Nous implémentons cette approche en utilisant une politique ordinale avec 17 classes par dimension pour résoudre divers environnements de référence en RL, notamment \texttt{Mujoco} (Todorov \textit{et al.}, 2012). L’optimisation de la politique est réalisée via \textsf{PPO}, avec l’implémentation des actions continues de Huang \textit{et al.}, 2022, et les mêmes hyperparamètres.

Nous utilisons le réseau de neurones implémenté pour l'algorithme \textsf{PPO} en actions continues dans Huang \textit{et al.}, 2022, basé sur un MLP à deux couches pour paramétrer $g_{\omega}(\cdot)$, ainsi qu’un écart-type à estimer par dimension d’action, indépendant de l’état. Les expériences sont menées dans les environnements du Tableau \ref{tab:recap_env}, avec un taux d’apprentissage de $3 \cdot10^{-4}$ et un facteur d’actualisation $\gamma=0.99$ pour \textsf{PPO} en actions continues et \textsf{PPO} ordinal. Les résultats montrent que la politique ordinale atteint des performances équivalentes, voire légèrement supérieures, à celles de la politique continue dans tous les environnements testés.

\begin{table}[h]
    \centering
{\footnotesize
\begin{tabular}{l c c| l c c} \toprule
   \textit{Environnement} & $|\mathcal{S}|$ & $|\mathcal{A}|$ & \textit{Environnement} & $|\mathcal{S}|$ & $|\mathcal{A}|$ \\
    \midrule
    \texttt{Ant-v4} & 105 & 8  & \texttt{BipedalWalker-v3} & 24 & 4  \\
    \texttt{HalfCheetah-v4} & 17 & 6  & \texttt{Hopper-v4} & 11 & 3 \\
    \texttt{Humanoid-v4} & 348 & 17 & \texttt{InvertedDoublePendulum-v4} & 9 & 1  \\
    \texttt{Pusher-v4} & 23 & 7  & \texttt{Walker2d-v4} & 17 & 6 \\
    \bottomrule
\end{tabular}
\caption{Environnements à actions continues où des actions ordinales ont été utilisées via la discrétisation.}
\label{tab:recap_env}
}%
\end{table}

\begin{figure}[h!]
    \centering
    \includegraphics[width=0.8\linewidth]{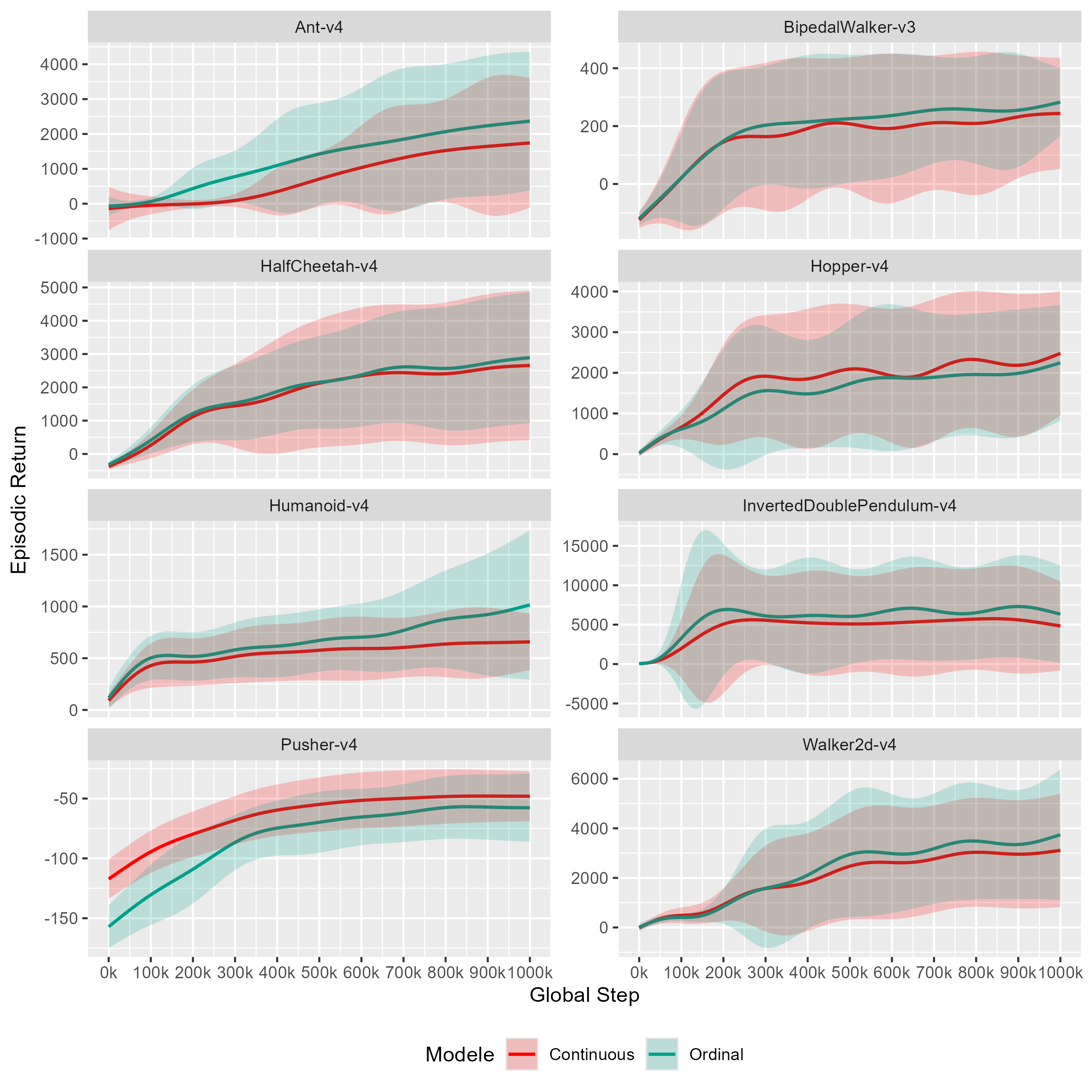}
    \caption{Courbes d'apprentissage pour différentes politiques sur divers environnements. Moyenne et intervalle de confiance à 95\%, ajustés à l’aide d’un modèle GAM gaussien à échelle localisée sur des données générées avec cinq tirages aléatoires par environnement.}
    \label{fig:learning_curves}
\end{figure}

\section{Conclusion}


L’essor des dispositifs portables hautement configurables ouvre la voie à des modèles adaptatifs répondant aux besoins individuels. Le RL fournit un cadre puissant pour ce type de problèmes, mais son déploiement en conditions réelles nécessite de prendre en compte les contraintes spécifiques à chaque application. Lorsqu’un système contrôle un paramètre ordonné, l’adoption d’une politique ordinale garantit une robustesse structurelle : quelles que soient les mises à jour, la politique conserve la hiérarchie des actions, assurant ainsi une meilleure cohérence.


Enfin, la discrétisation des actions continues demeure un enjeu ouvert. Dans la Section \ref{sec:discretization}, nous avons fixé arbitrairement le nombre de classes et la méthode de discrétisation, mais une approche plus adaptative pourrait être explorée. Par exemple, au fil de l’apprentissage, de nouvelles classes pourraient être ajoutées dynamiquement en conservant le prédicteur appris $g_{\omega}$ et en introduisant de nouveaux seuils, optimisant ainsi la granularité de la politique.


\section*{Bibliographie}




\noindent Agarwal, A., Kakade, S. M., Lee, J. D., et Mahajan, G. (2020) Optimality and approximation with policy gradient methods in markov decision processes. In \textit{Conference on Learning Theory}, PMLR, pp.~64-66.





\noindent Huang, S., Dossa, R.F.J., Ye, C., Braga, J., Chakraborty, D., Mehta, K., Araújo, J.G.: Cleanrl: High-quality single-file implementations of deep reinforcement learning algorithms. Journal of Machine Learning Research 23(274), 1–18 (2022),

\noindent Kakade, S. M. (2001). A natural policy gradient. Advances in neural information processing systems, 14.





\noindent Schulman, J., Levine, S., Abbeel, P., Jordan, M. et Moritz, P. (2015) Trust Region Policy Optimization. \textit{Proceedings of the 32nd International Conference on Machine Learning}, PMLR

\noindent Schulman, J., Wolski, F., Dhariwal, P., Radford, A., et Klimov, O. (2017). Proximal policy optimization algorithms. arXiv preprint \texttt{arXiv:1707.06347}.



\noindent Sutton, R.S., Barto, A.G.: Reinforcement Learning: An Introduction. A Bradford
Book, Cambridge, MA, USA (2018)


\noindent Todorov, E., Erez, T., Tassa, Y.: Mujoco: A physics engine for model-based control. In: 2012 IEEE/RSJ International Conference on Intelligent Robots and Systems. pp. 5026–5033. IEEE (2012).


\end{document}